\documentclass{amia}
\usepackage{graphicx}
\usepackage[labelfont=bf]{caption}
\usepackage[superscript,nomove]{cite}
\usepackage{color}
\usepackage{tabularx}
\usepackage{multirow}
\usepackage{multicol}

\usepackage{comment}

\begin{document}

\title{Extracting Daily Dosage from Medication Instructions in EHRs: \break An Automated Approach and Lessons Learned}

\author{Diwakar Mahajan, MS, Jennifer J. Liang, MD, Ching-Huei Tsou, PhD}

\institutes{
    IBM T.J. Watson Research Center, Yorktown Heights, NY\\
}

\maketitle
\noindent{\bf Abstract}

\textit{Medication timelines have been shown to be effective in helping physicians visualize complex patient medication information. A key feature in many such designs is a longitudinal representation of a medication's daily dosage and its changes over time. However, daily dosage as a discrete value is generally not provided and needs to be derived from free text instructions (Sig). Existing works in daily dosage extraction are narrow in scope, targeting dosage extraction for a single drug from clinical notes. Here, we present an automated approach to calculate daily dosage for all medications, combining deep learning-based named entity extractor with lexicon dictionaries and regular expressions, achieving 0.98 precision and 0.95 recall on an expert-generated dataset of 1,000 Sigs. We also analyze our expert-generated dataset, discuss the challenges in understanding the complex information contained in Sigs, and provide insights to guide future work in the general-purpose daily dosage calculation task.}

\section*{Introduction}

Medication histories are an integral part of a patient's medical history and have substantial impact in the patient's clinical care process. An accurate medication history helps providers detect potential medication-related pathologies, assess efficacy of current treatments, determine necessary treatment changes going forward, and avoid prescription errors (e.g. restarting previously discontinued medication, prescribing wrong dose). 
With the increasing prevalence of polypharmacy, especially in the elderly and patients with multiple comorbidities, understanding the patient's complete medication history becomes a complex and challenging task using today's electronic health record (EHR) systems, which typically records such information in data tables or clinical notes. Previous studies have reported anywhere from 10\% to 90\% of patients taking 5 or more medications depending on the study population\cite{refKhezrian}, with excessive polypharmacy (i.e. taking 10 or more medications) reported in anywhere from 18\% to 39\% of patients\cite{refSalvi, refKim}.

\begin{figure}[h!]
\centering
\includegraphics[scale=0.7]{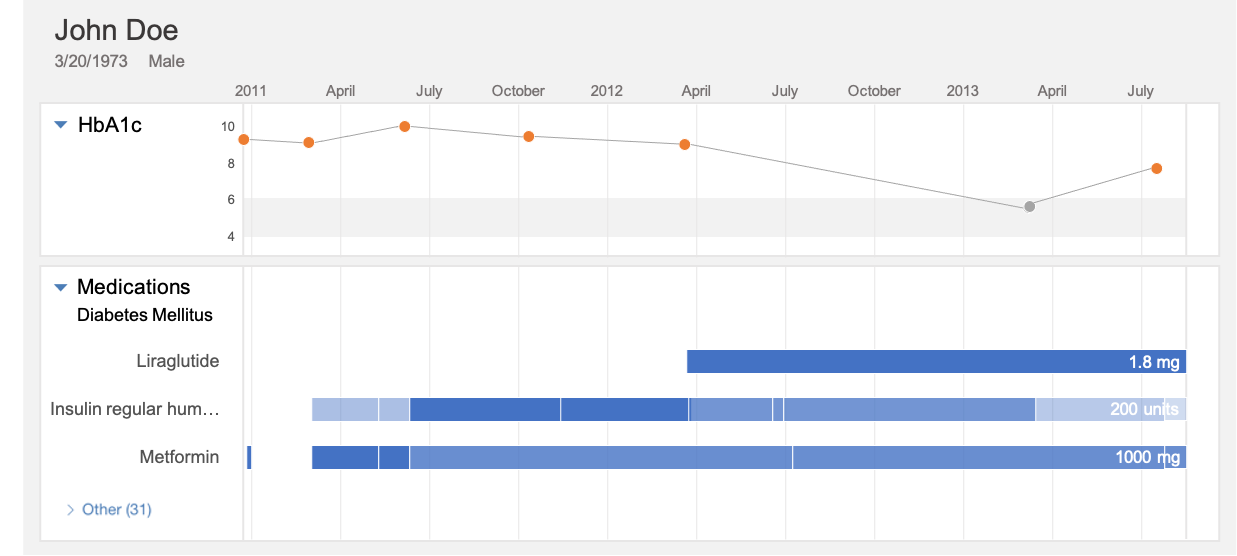}
\caption{Medication timeline prototype for diabetes mellitus. Intensity of medication bar corresponds to daily dosage, where lighter indicates lower dosage and darker indicates higher dosage.}
\label{fig:medtimeline}
\end{figure}

These long and complex medication lists add to the cognitive load on providers in their daily workflow, making it easier to miss important information with potential negative consequences. Graphical timelines have long been proposed as an effective way to visualize patient medication information, with the ability to capture and represent the historical and longitudinal nature of EHR data. Plaisant et al.\cite{refLifelines} presented Lifelines in 1996, which included both a broad and detailed timeline view of medication details. More recently, Belden et al.\cite{refBelden} developed a prototype medication timeline design that demonstrated improved physician performance in 5 common medication-related tasks on a pilot evaluation. Figure \ref{fig:medtimeline} presents our version of a medication timeline in a diabetes mellitus (DM)-specific view of the patient record, which shares many of the same features as the designs proposed by previous researchers.

As demonstrated in Figure \ref{fig:medtimeline}, such a view allows the user to more easily perform medication-related tasks, such as identifying new prescriptions or dosage changes in a given time interval, which are cumbersome to do otherwise\cite{refBelden}. Furthermore, by viewing medication timelines together with timelines of other clinical information, such as relevant laboratory results, physicians are able to identify potential reasons for changes in patient status. For example, in Figure \ref{fig:medtimeline}, the decrease in HbA1c, indicating improvement in the patient's DM, may be due to the introduction of a new medication liraglutide to the patient's DM regimen; there is also a later increase in HbA1c (i.e. worsening DM) that may be attributed to a decrease in insulin prescribed at an earlier visit. 

A key feature in many medication timeline designs, including ours, is the use of the intensity of the timeline bar to indicate the dosage of medication prescribed. However, dosage as a discrete quantity is generally not provided and requires understanding the free text Sig field in combination with other details of the medication itself in order to calculate it. To address this gap, we developed an automated system to calculate daily dosage using information provided in EHR structured medication data.



Previous studies on daily dosage calculation have focused on clinical notes, with each study targeting dosage extraction for a single drug of interest. Specifically, two past works by Xu et al. targeted dosage extraction for Tacrolimus\cite{refTacrolimus} and Warfarin\cite{refWarfarin} using a semantic grammar, lexicon and rule-based system\cite{medex}. While these systems show high performance, these studies are limited to a single medication and therefore restricted in Sig language variability. Further exploration is required to demonstrate the effectiveness of such approaches to extract daily dosage for all medications towards generating a complete timeline visualization of a patient's medication history. 
In our work we expand our scope to target all medications, 
thereby introducing more variability and complexity to the task. To the best of our knowledge this is the first work targeting general daily dosage calculation for all medications. 

The Sig for a medication prescription provides specific directions to the patient on how to take the medication. These directions typically include information on how often to take the medication and how much of the medication to take each time (e.g. \textit{`Take 1 tablet twice daily'}), and may sometimes include more detailed instructions such as when to take the medication, how to take the medication, why the patient is taking the medication, etc. The Sig field can be populated in various ways, such as drop-downs, templates, free text, or some combination of the above, resulting in numerous textual variations in the actual Sig language. A prior study of electronic prescriptions in ambulatory care settings found 832 permutations of the Sig concept of `\textit{Take 1 tablet by mouth once daily.}'\cite{refSigQuality}. Due to the free text nature of Sig text, sometimes confusing Sigs are generated that are incomplete, ambiguous, or contradictory leading to difficulty in interpretation\cite{refSigQuality}.
Because of the multitude of prescriptions available in medicine and the variable language and quality of the Sig field, it is important for a system to (1) be extensible to a large variety of medication prescriptions, (2) only return a daily dosage value when there is a sufficient level of confidence, and (3) be able to flag prescriptions that may contain confusing or ambiguous Sigs. To address the real world need and system requirements listed above, in this paper, we:
\vspace{-2mm}
\begin{enumerate}
\addtolength\itemsep{-2mm}
    \item Characterize the complexities in medication prescriptions and the challenges in calculating a daily dosage,
    \item Describe how we've defined our task to apply to a wide variation of medications and Sig language,
    \item Present our hybrid system leveraging a publicly available dataset to overcome the lack of ground truth,
    \item Validate our system on an expert-generated dataset.
\end{enumerate}
\vspace{-2mm}







\section*{Materials and Methods}



\textbf{\textit{Data and Annotation Process}}

We sampled 1,000 medication orders from 427 patients within a large multidisciplinary medical center for use in ground truth generation for system evaluation. This dataset contained 296 unique medications, 15 different routes of administration, and 68 different formulations.
Annotators were provided with the Sig text along with the accompanying medication strength (i.e. the amount of drug in a given dosage form, e.g. `500 mg') to calculate the daily dosage. Table \ref{table:gt} provides some sample annotations.

\vspace{2mm}
\begin{table}[h!]
\centering
\caption{Sample ground truth (GT) annotations for daily dosage.}
  \begin{tabular}{|l|l|l|l|l|} \hline
    \multicolumn{2}{|c|}{\textbf{Medication Order Information}} & \multicolumn{2}{c|}{\textbf{Daily Dosage GT}} & \multirow{2}{1.7cm}{\centering\textbf{Comments}}\\ \cline{1-4}
    \textbf{Sig} & \textbf{Strength} & \textbf{Max} & \textbf{Min} &  \\ \hline
    1/2 tab bid & 2 mg & 2 mg & -- & -- \\ \hline
    
    Take one(1) inhalation twice daily & 250-50 mcg/dose & 500-100 mcg & -- & Multiple ingredients \\ \hline
    one to two tablets daily & 7.5 mg & 15 mg & 7.5 mg & Sig indicates a range \\ \hline
    1 tablet every day except on Monday at 4pm & 5 mg & 4.29 mg & -- & Non-daily dose \\ \hline
  \end{tabular}
\label{table:gt}
\end{table}

In the case of combination medications, where a single formulation contains multiple ingredients (e.g. strength = \textit{`300-30 mg'} indicates a medication with two ingredients, one with strength = `300 mg' and the other with strength = `30 mg'), daily dosage was calculated separately for each ingredient.
When a range is indicated in the Sig text (e.g. \textit{`1-2 tabs every 4-6 hours'}), annotators were asked to provide the minimum and maximum daily dosage the patient could take. 
For example, given the Sig text \textit{`take 1 tab po TID- QID'} for medication with strength \textit{`10-325 mg'}, the annotator would provide the following as ground truth: Min Daily Dosage = `\textit{`30-975 mg'}; Max Daily Dosage = \textit{`40-1300 mg'}.

Sometimes a medication is prescribed to be taken regularly but not necessarily daily (e.g. every other day), as is often the case when the unit of medication does not come in the desired dose per day (e.g. physician wants a daily dose of 5 mg but the medication only comes in 10 mg tablets). Since our purpose in extracting daily dosage is to plot it on a timeline, for these cases annotators were asked to calculate the average daily dosage if the patient was taking at least one dose over the course of a week. For example, the annotator would annotate a daily dosage for the Sig \textit{`1 tab po q week'} by dividing the total weekly dose over 7 days. When the frequency of dosing is less than once a week (e.g. \textit{`1000mcg IM monthly'}), annotators were asked to mark this as `no daily dosage'. Similarly, if the annotator could not calculate a daily dosage for any other reason, they would indicate so and provide the reason why.


To better understand the complexities in Sig expressions and the challenges of annotating daily dosage extraction task, we present an analysis of the ground truth generated by medical experts in the Results section. We report on the number of prescriptions where human experts can and cannot determine a daily dosage. We further characterize those occurrences based on (1) medication attributes provided by the structured medication entry and (2) qualitative observations noted by our medical experts during the annotation process.

\textbf{\textit{System Description}}

\begin{figure}[h!]
\centering
\includegraphics[scale=0.5]{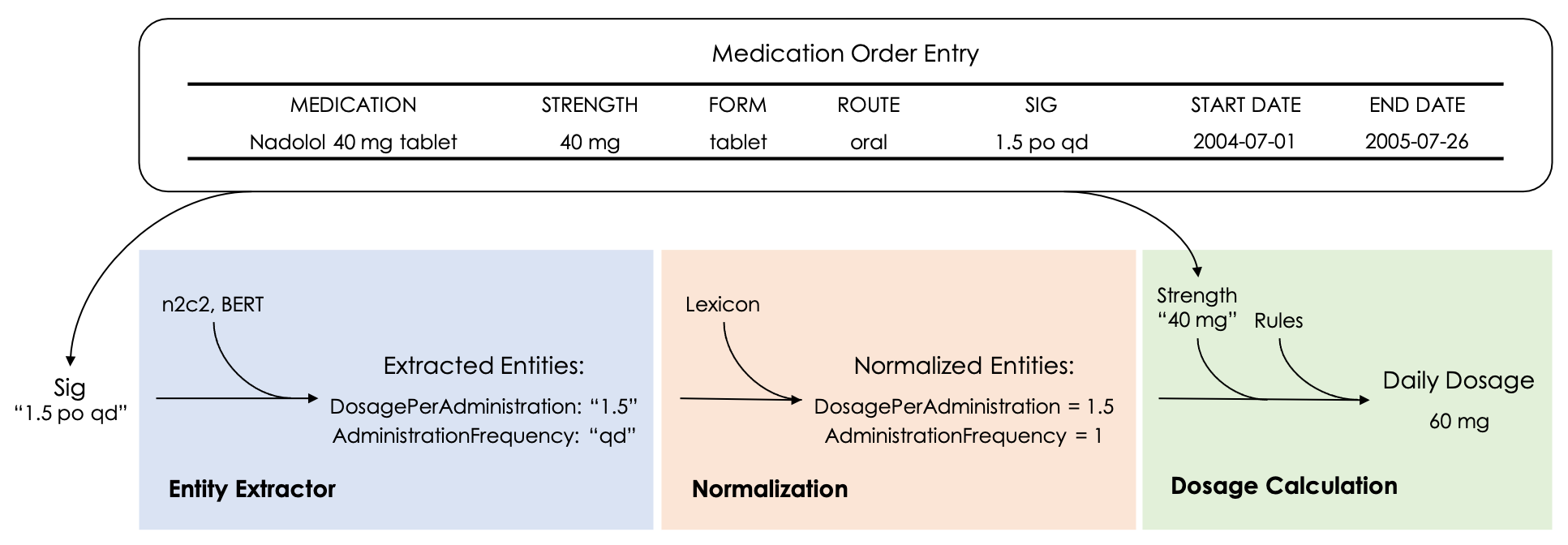}
\caption{Overview of the daily dosage calculating system.}
\label{fig:systemoverview}
\end{figure}

We define the daily dosage calculation task as follows: given a Sig and accompanying medication strength, calculate the minimum and maximum prescribed daily dosage for the medication. To accomplish this, we undergo a two step process to (1) extract entities of interest and their spans from the Sig and (2) normalize extracted entities and combine the normalized values to calculate the daily dosage value. We achieve the first step using an entity extractor and the second step by a normalization and dosage calculation module. Both are explained in the following sections. An overview of the system is presented in Figure \ref{fig:systemoverview}.

\textit{Entity Extraction Module}

We define two entities of interest for extraction by the entity extraction module: DosagePerAdministration (DA) and AdministrationFrequency (AF).
\vspace{-2mm}
\begin{itemize}
\addtolength\itemsep{-2mm}
    \item DA is the dosage or the amount of medication that should be taken per administration of the medication. It is usually expressed either as a number (e.g. \textit{`2'}), a number followed by the form (e.g. \textit{`1 tablet'}), or a number followed by the unit (e.g. \textit{`25 mg'}). 
    \item AF is the number of administrations of the medication that are executed over a period of time. It is usually expressed as frequency of the medication (e.g. \textit{`two times a day', `once a week'}). Note that the time frame referenced by AF can vary depending on the Sig. 
\end{itemize}
\vspace{-2mm}

Relevant to the task of daily dosage calculation, considerable work has been done in the area of clinical entity extraction. Previous studies have created annotations to extract medications and related attributes (e.g. dosage, strength, form, route, etc.) from clinical notes\cite{med2010,n2c2_1,n2c2_2}. 
For automating the task of medication information extraction from clinical text, previous works have employed various flavors of Bidirectional Encoder Representations from Transformers (BERT\cite{bert})-based language models\cite{bluebert}, with the models pretrained on relevant domain (e.g. biomedical domain - BioBERT\cite{biobert}, clinical domain  - ClinicalBERT\cite{clibert}) showing state-of-the-art performance\cite{biobert,sotaadr,clinicalent}.

To overcome the lack of data for entity extraction in our dataset, we utilized the publicly available n2c2 2018 Adverse Drug Events and Medication Extraction in EHRs dataset\cite{n2c2_1,n2c2_2}, which contains 505 clinical notes labeled with entities such as Drug, Dosage, Strength, Frequency, etc. As there is a considerable overlap in definition, we combine n2c2 entities using some simple rules to form annotations for our entities: DA and AF (Table \ref{table:transferentities}).

\vspace{2mm}
\begin{table}[h]
\centering
\label{table:transferentities}
\caption{Proxy ground truth for DosagePerAdministration (DA) and AdministrationFrequency (AF) entities.}
\begin{tabular}{|l|l|l|l|}
\hline
 \textbf{Entity Type}  & \textbf{Overlapped n2c2 entity}  & \textbf{Examples}  & \textbf{Size}\\ 
  \hline
DosagePerAdministration  & (n2c2\_Dosage or n2c2\_Strength) + n2c2\_Form  & 1-2 tabs, 3 puffs & 6,902\\
\hline
AdministrationFrequency  & n2c2\_Frequency & daily, B.I.D. & 10,362\\
\hline
\end{tabular}
\label{table:transferentities}
\end{table}

We created train, development and test splits (75/5/20) on this proxy ground truth and trained transformer-based language models to automate the entity extraction task. We employed ClinicalBERT\cite{clibert} in our experiments for extracting daily dosage relevant clinical entities.
In this process, we first obtained contextualized word embeddings from ClinicalBERT, and then used a Bidirectional Long Short-Term Memory\cite{bilstm} of size 100, 0.5 dropout and an additional fully connected layer of size 40 with soft-max activation to train a sequence-to-sequence model that outputs the entities. We used the transformers package\cite{transformers} to tune our models with the train and development splits, and present our results on the test split.

\textit{Normalization \& Dosage Calculation Module}

After the entity extractor has identified spans for the various entities, the extracted spans are then normalized to their corresponding numerical values using a pre-built lexicon. 
A simple set of rules, presented in Table \ref{table:rules}, was then applied to combine these normalized values to calculate the final daily dosage value. 

\vspace{2mm}
\begin{table}[h]\centering
\caption{Daily dosage identification and calculation examples.}
\begin{tabular}{|>{\raggedright\arraybackslash}m{3.5cm}|>{\raggedright\arraybackslash}m{2.7cm}|>{\raggedright\arraybackslash}m{2.3cm}|>{\centering\arraybackslash}m{2.7cm}|>{\centering\arraybackslash}m{2cm}|}
\hline
\multirow{2}{*}{\textbf{Medication Order}}
 & \multirow{2}{*}{\textbf{Entity Extraction}} & \multirow{2}{*}{\textbf{Normalization}}
 & \multicolumn{2}{c|}{\textbf{Daily Dosage Calculation}}
\\ 
 \cline{4-5}
 
& & & \centering\textbf{Rule Applied} & \textbf{Daily Dosage} \\ 
 \hline
\textbf{Sig:} Take two tablets twice daily  \break \textbf{Strength:} 50mg &
\textbf{DA}: two tablets \break\textbf{AF}: twice daily & 
\break\textbf{DA}: 2 \break\textbf{AF}: 2
 &
DA * AF * Strength 
& 200 mg\\
\hline
\multirow{2}{3.5cm}{\textbf{Sig:} Take one tab in am and two tabs in pm \break \textbf{Strength:} 50mg} 
& 
\textbf{DA1}: one tab
\break\textbf{AF1}: am
& 
\textbf{DA1}:1 \break\textbf{AF1}: 1
&
\multirow{2}{2.7cm}{(DA1 * AF1 * Strength)
 + (DA2 * AF2 * Strength)}
&
\multirow{2}{*}{150 mg}\\
\cline{2-3}
&
\textbf{DA2}: two tabs
\break\textbf{AF2}: pm
&
\textbf{DA2}: 1 \break\textbf{AF2}: 2 
& & \\ \hline
\end{tabular}
\label{table:rules}
\end{table}

\textbf{\textit{Evaluation Metric}}

We report two sets of evaluations, (1) evaluation of the entity extraction module on the proxy ground truth obtained from n2c2 dataset and (2) evaluation of the end-to-end system on expert-generated ground truth. We report on the precision, recall, and F-score for both tasks.
For the entity extraction evaluation, we performed a strict evaluation, i.e. exact matching of both span and type of entity. For the end-to-end evaluation, the system was judged to be correct when it returns the same minimum and maximum daily dosage value and unit of measurement as the human expert. For medications with multiple ingredients, the system must correctly return the daily dosage value for each individual ingredient. If the system returns a daily dosage value when the human expert has indicated that a daily dosage value could not be determined, it is counted against the system as a false positive. Similarly, if the system is unable to return a daily dosage when one is provided in the ground truth, it is counted as a false negative. In the case that the system returns a daily dosage value that differs from that in the ground truth, it is counted as both a false positive and false negative.


\section*{Results}

\textbf{\textit{Ground Truth Analysis}}

\begin{figure}[h!]
\centering
\includegraphics[scale=0.55]{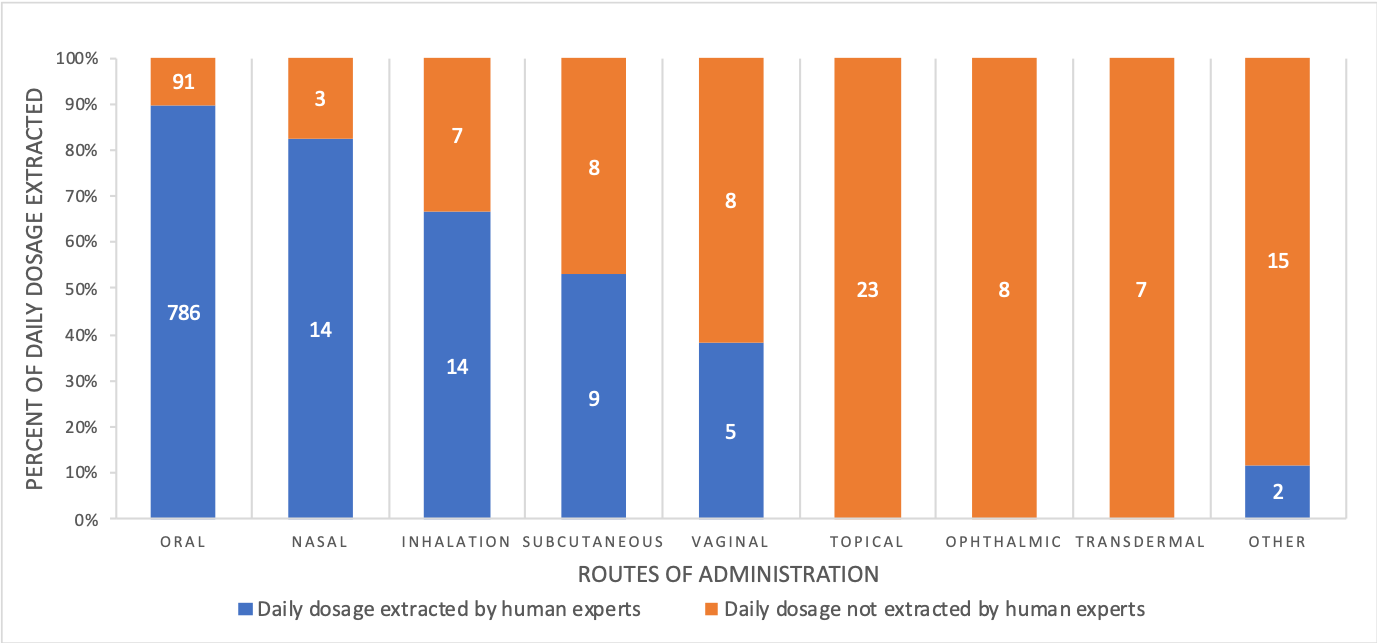}
\caption{Percent of medication orders annotated with daily dosage by human experts for different routes.}
\label{fig:route}
\end{figure}

Analysis of the human expert-generated ground truth reveals that human experts are only able to provide daily dosage in 83\% of the data, as seen in Figure \ref{fig:route}. Within this subset of data with expert-provided daily dosage values, 94.7\% are medications delivered orally, 1.7\% through inhalation, 1.7\% nasal, 1.1\% subcutaneous, and the remaining 0.8\% consist of other routes of administration. For the remaining 17\% of cases where human experts did not provide a daily dosage value, we explore the reasons behind this finding with specific examples in the Discussion section.

\vspace{2mm}

Of the 830 medications where human experts provided a daily dosage value, 14.7\% had multiple ingredients and 8.4\% had Sigs that included ranges, indicating multiple daily dosage values for a given prescription (e.g. minimum / maximum, daily dose per ingredient), and adding complexity to the Sig analysis. 

\textbf{\textit{System Performance}}

Table \ref{table:DLentityextractor} presents the results of the transformer-based entity extractor evaluated on the proxy ground truth test set based on n2c2 2018 dataset. 
\vspace{2mm}
\begin{table}[h!]
\centering
\caption{Precision, recall, and F-score of entity extractor.}
  \begin{tabular}{|l|c|c|c|}
  \hline
    \textbf{Entity Type}  & \textbf{Precision}  & \textbf{Recall} & \textbf{F-score} \\ \hline
      DosagePerAdminstration  & 0.989  & 0.985 & 0.987 \\ \hline
      AdministrationFrequency  & 0.990 & 0.981 &0.985 \\ \hline 
  \end{tabular}
\label{table:DLentityextractor}
\end{table}

Table \ref{table:evaluationAll} presents the results of our end-to-end dosage calculation system. 
Our system achieved 0.98 precision, 0.95 recall, 0.96 F-score and 96.2\% accuracy on the evaluation dataset of 1,000 Sigs. Another way to understand these results is by comparison to daily dosage as provided by human experts in the ground truth. In our evaluation dataset, human experts provided a daily dosage in 83\% of the data, and is 100\% correct when a daily dosage value is returned. Whereas, our system returned a daily dosage in 81.6\% of the data, and is 98.2\% correct when returned.
\vspace{2mm}
\begin{table}[h]
\centering
\caption{System evaluation results on all 1,000 prescriptions.}
  \begin{tabular}{|c|>{\centering\arraybackslash}m{2.5cm}|>{\centering\arraybackslash}m{3.5cm}|>{\centering\arraybackslash}m{3.5cm}|>{\centering\arraybackslash}m{2.5cm}|}
  \hline
    \multicolumn{2}{|c|}{\multirow{2}{*}{ }} &  \multicolumn{3}{c|}{\textbf{System}} \\ \cline{3-5}
    \multicolumn{2}{|c|}{}  & \textbf{Daily dosage extracted and CORRECT}  & \textbf{Daily dosage extracted but INCORRECT} & \textbf{Daily dosage NOT extracted}   \\ \hline
    \multirow{3}{1.5cm}{\centering\textbf{Human Expert}} & \textbf{Daily dosage extracted}  & 800 & 7 & 23 \\ \cline{2-5}
    &\textbf{Daily dosage NOT extracted}  & not applicable & 8 & 162 \\ \hline
  \end{tabular}
\label{table:evaluationAll}
\end{table}

\section*{Discussion}

We described our automated system for the general-purpose daily dosage calculation task on all medications, and presented our evaluation on an expert-generated dataset of 1,000 Sigs. The high performance of our system suggests that even with the additional variability introduced by a wider range of medications, the proposed approach is sufficient to cover most cases. 
However, analysis of our expert-generated dataset reveal that (1) the system can further benefit from different normalization modules for medications with different routes of administration, and (2) there is need for future work to understand Sig information beyond daily dosage (e.g. PRN, time-limited dosing, specific indications, cases where daily dosage not meaningful). We discuss these findings in more detail in the subsequent sections.



\textbf{\textit{Analysis of Expert-Generated Ground Truth}}

As observed in Figure \ref{fig:route}, human experts did not provide a daily dosage value in 17\% of prescriptions. This 17\% consists mostly of medications delivered either by oral, topical, subcutaneous, vaginal, or ophthalmic routes of administration. In the following section we discuss these cases as organized by the medication's routes of administration.

\begin{figure}[h!]
\centering
\includegraphics[scale=0.5]{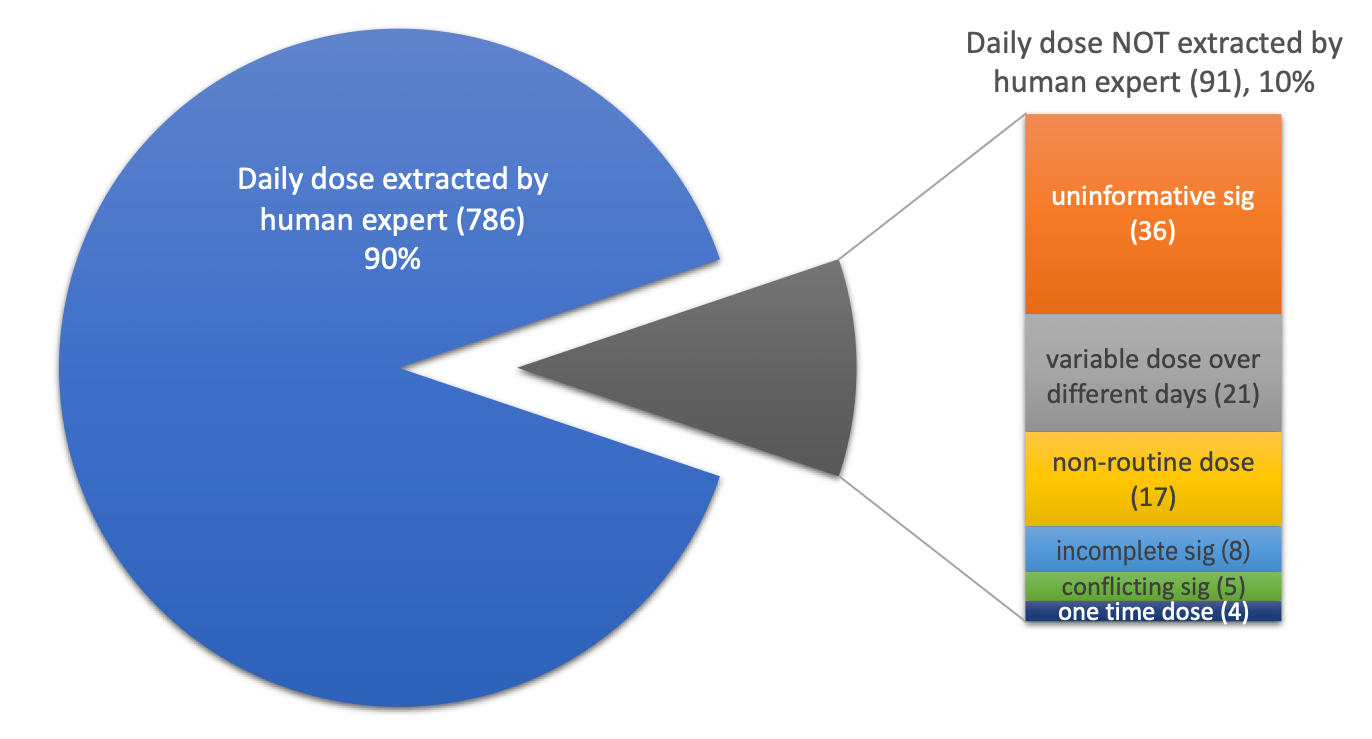}
\caption{Analysis of human expert annotated daily dosage ground truth for 877 oral medications.}
\label{fig:gt-analysis-oral}
\end{figure}

\textit{Oral.} For oral medications, we identified three major categories of reasons for why human experts did not provide a daily dosage value:
\vspace{-2mm}
\begin{enumerate}
\addtolength\itemsep{-2mm}
    \item Need more information, where Sig alone is insufficient to determine daily dose, including Sigs that are lacking in detail (uninformative sig), missing information (incomplete sig), or presenting conflicting instructions.
    \item Variable dose over different days, where the daily dosage varies from day to day.
    \item Daily dose not meaningful, where the medication is prescribed either for use under a specific set of circumstances (non-routine dose), or as a one time single administration (one time dose).
\end{enumerate}
\vspace{-2mm}
Figure \ref{fig:gt-analysis-oral} is an analysis of oral medication prescriptions reviewed by human experts, and shows the number of oral prescriptions where the human experts calculated a daily dosage, the number of prescriptions where they did not provide a daily dosage, and the reasons why.

The largest category of oral medication orders where experts did not provide a daily dosage were cases where more information beyond the Sig was needed to determine the daily dosage value, either because the Sig was uninformative, missing key information, or contained ambiguous or conflicting instructions that need clarification. In the case of uninformative Sigs, the actual patient instructions may be written elsewhere. However, Sigs with either missing, ambiguous, or conflicting instructions reflect a Sig quality issue, which has been studied in more detail by Yang et al\cite{refSigQuality}. The second major category were Sigs that indicate varying dosing over different days. A common example is a tapering dose, where the patient is instructed to start taking the prescription at a specific dose, then gradually reduce the dose over a certain number of days until eventually the medication is discontinued. Since the daily dosage value changes depending on the specific day, our current representation of a single daily dosage value is insufficient to capture these cases of varying dosages. Lastly, in some cases the drug was not taken regularly enough for a daily dosage value to be meaningful. Common examples are prescriptions with instructions to only take the medication in a specific scenario, such as \textit{`30-60 minutes before sexual intercourse'}, \textit{`one hour prior to procedure'}, or \textit{`at onset of headache'}. Table \ref{table:exampleNoDD} shows some examples of Sigs for oral medications that fall under each of these categories.

\vspace{2mm}

\begin{table}[h]
\centering
\caption{Examples of Sigs for oral medications where human experts declined to provide a daily dosage value.}
  \begin{tabular}{|m{8.7cm}|m{6.8cm}|}
  \hline
    \textbf{Sig Text}  & \textbf{Comments} \\ \hline
    Take as directed.  & Need more information - uninformative Sig \\ \hline
    Take 1 tablet by mouth.  & Need more information - missing frequency \\ \hline
    Take 0.25 tablets by mouth once daily. TAKE ONE HALF (0.5) OF A  TABLET DAILY.  & Need more information - conflicting instructions \\ \hline
    
    Take one and half (1.5) tablets twice daily (weeks 1-4); Take one and half (1.5) tablet in AM and two(2) tablet in PM (week 5); Take two(2) tablets twice daily from week 6 onwards.  & Variable dose over different days \\ \hline
    Take 6 tab day1, 5 tab day 2, 4 tab day3 , 3 tab day 4, 2 tab day 5, 1 tab day 6.  &  Variable dose over different days \\ \hline
    
    Take 4 pills by mouth one hour prior to the procedure.  & Daily dosage not meaningful - non-routine dose \\ \hline
    Take 1 tablet by mouth one time only.  & Daily dosage not meaningful - one time dose \\ \hline

  \end{tabular}
\label{table:exampleNoDD}
\end{table}

\textit{Topical.} For most other routes, the reason why human experts declined to provide a daily dosage is often closely related to the way these medications are formulated or delivered. For example, topical medications usually come in the form of creams, gels, ointments, lotions or similar formulations, which are not easily quantified by a specific measurable amount. Therefore, Sigs are more generally written, such as \textit{`Apply to affected area twice daily'}, and do not translate to a specific daily dosage value. This observation also applies to some medications delivered by other routes of administrations, such as ophthalmic ointments and rectal and vaginal creams.

\textit{Ophthalmic.} Some ophthalmic medications are formulated as solutions or suspensions, which are measurable in milliliters. However, ophthalmic solutions or suspensions are typically delivered in the form of drops, such as \textit{`Use 1 Drop in the left eye twice daily'}. Since the volume of a drop is not well-defined, there is no standard normalization of this unit to convert these types of Sig expressions into a daily dosage value.

\textit{Subcutaneous.} In the case of medications delivered through injections, either subcutaneously, intramuscularly, or intravenously, a common reason cited by human experts for why a daily dosage value could not be extracted is because the medication was indicated as a one time dose, for example \textit{`Inject 0.65 mL subcutaneously one time only for 1 dose'}. This is likely because such medications are more often delivered in a medical setting by an experienced medical professional familiar with delivering injections. 




\textbf{\textit{Error Analysis on the System Performance}}

To better understand and improve our system, we reviewed all cases where the system erred, either because it (1) returned a daily dosage different from what is in the ground truth, (2) returned a daily dosage where human expert did not, or (3) failed to return a daily dosage when one was provided in the ground truth. Due to the patient safety implications of returning an incorrect daily dosage, we developed our system to have very high precision. However, there were still a few cases where our system returned an incorrect daily dosage. We identified the following four main categories of errors, with some examples shown in Table \ref{table:exampleErrors}:
\vspace{-2mm}
\begin{enumerate}
\addtolength\itemsep{-2mm}
    \item Multiple dosage expressions - 20\%
    \item New expressions (not understood by system) - 20\%
    \item Extra information (not considered by system) - 17.2\%
    \item Typo and/or misspellings - 11.4\%
\end{enumerate}
\vspace{-2mm}

\vspace{2mm}

\begin{table}[h!]
\centering
\caption{Examples of system errors.}
  \begin{tabular}{|m{8.5cm}|m{7cm}|}
  \hline
  \textbf{Sig Text}  & \textbf{Comments} \\ \hline
Take 1 tablet by mouth daily before lunch. take one full pill before dinner & Multiple dosage expressions - system could not add two expressions  \\ \hline
Take 2.5 mg by mouth once daily. Warfarin 5 mg only on Tues and Thurs, other days takes 2.5 mg. & Multiple dosage expressions - system unable to reconcile between the two \\ \hline

      alternates 5 mg and 7.5 mg daily & New expression `alternates' \\ \hline
    Take one tablet Mon-Wed-Thur-Sat  & New expression - `Mon-Wed-Thur-Sat'  \\ \hline
    Take 1-2 tablets by mouth every 6 hours as needed for Pain (max = 6 tabs/day).  & Extra info (daily limit) not considered by system \\ \hline

    2 po bid and maytake an extra 1/2 tab qd prn palpiations & Extra info (`extra') not considered by system \\ \hline
    1 talbet 5times/day  & Typo / misspelling \\ \hline

  \end{tabular}
\label{table:exampleErrors}
\end{table}

During error analysis, we observed various scenarios that result in a Sig containing multiple dosage expressions. Sometimes this is intentional, for example when the provider instructs the patient to take different doses at different times of day (e.g. \textit{`Take 1 tablet by mouth daily before lunch. take one full pill before dinner'}). For these cases of variable dosing throughout the day, the system attempts to understand this as a separate dosage expression for each administration, and then adds them up to calculate the daily dosage. However, depending on the language of the Sig, the system may not be able to construct full dosage expressions for each administration, and therefore opts to return null in cases of insufficient confidence, resulting in a false negative. Just as often though, multiple dosage expressions in a Sig is unintentional, likely due to user error in generating the Sig. We observed cases where the Sig contained duplicate instructions (e.g. \textit{`Take by mouth every 4 hours as needed for Cough. Take 1 teaspoon(s) as needed for cough every 4 hrs.'}) or conflicting instructions (e.g. \textit{`1 tablet daily Take one tablet every other day'}). 
Although the system was able to identify most cases of conflicting Sigs and returned null in those cases, it missed two instances where instead of returning null, it returned an incorrect daily dosage calculated based on only the first dosage expression.

Next, we observed a sizeable chunk of errors belonged to the new expressions category, where new phrases or words were encountered by the system. These include phrase like \textit{`alternate'} or specific frequency instructions like \textit{`Mon-Wed-Thurs-Sat'}. 
We also observed error cases where extra information in the Sig was not considered by the system, leading to an incorrect daily dosage. 
This extra information was usually either a maximum limit over a set time period that overrides the instructions in the rest of the Sig (e.g.  \textit{`max = 6 tabs/day'}, \textit{`Do not exceed 30 MG per day'}), or an exception to the instructions in the rest of the Sig (e.g. \textit{`2 po bid and maytake an extra 1/2 tab qd prn palpiations'}). For these cases, our system correctly parsed the first part of the Sig and returned a daily dosage value based on that, but did not consider the extra information to revise the daily dosage as necessary. Finally, we observed certain errors due to the system's failure to understand or parse typos or misspellings, such as \textit{`talbet'}.



\textbf{\textit{Future work}}

For future work, we plan to explore three broad ways to improve our current system. First, during this study we observed additional information imparted by Sig text not captured by a single daily dosage value (e.g. PRN, time-limited dosing, specific indications), and that certain types of medications have certain patterns of Sig text (e.g. tapering doses for steroids, time-limited administration for antibiotics). Therefore, we plan to surface these additional Sig characteristics, and leverage medication-specific information pulled from RxNorm to explore ways to further improve our system. Second, as observed by our analysis of the expert-generated ground truth, there are certain categories of Sigs where a daily dosage value should not be returned. Instead of simply returning null in these cases, we plan to surface the reason for the null value using those expert-identified categories, namely, `need more information' (e.g. uninformative, incomplete, or conflicting Sigs), `variable dosing over different days', and `daily dosage not meaningful' (e.g. non-routine or one time dose). Third, as the normalization component of our system remains rule-based and reliant on a curated lexicon, we aim to (1) explore seq2seq systems to normalize the entities or (2) develop an automated process to supplement this lexicon and calculate the final daily dosage value, with the end goal of developing an end-to-end learned system for daily dosage extraction.

\section*{Conclusion}
Understanding a patient's medication history is essential for physicians to provide appropriate treatment recommendations. A medication's prescribed daily dosage is a key element of the medication history and used in many medication timeline designs; however, it is generally not provided as a discrete quantity and needs to be derived from free text medication instructions. 
This work is the first to generalize the task of daily dosage extraction on all medications. We described our hybrid system for calculating daily dosage on all medications that consists of a deep learning-based entity extraction module and normalization and dosage calculation module, and demonstrated its effectiveness against an expert-generated dataset. We also presented an analysis of the expert-generated dataset that revealed certain shared characteristics among medications sharing the same route or similar formulations, and provided insights into additional, more nuanced information contained in Sigs. We hope this research will help realize many medication timeline designs and ultimately reduce the cognitive burden on healthcare providers, as well as support other research efforts requiring detailed dosage information, such as pharmacodynamic, pharmacokinetic and pharmacogenomic studies.

\makeatletter\centering
\renewcommand{\@biblabel}[1]{\hfill #1.}
\makeatother

\bibliographystyle{unsrt}

\end{document}